# Nonuniform dynamic discretization in hybrid networks

**Alexander V. Kozlov**
Department of Computer Science
Stanford University
Stanford, CA 94305-9010
e-mail: *alexvk@cs.stanford.edu*

**Daphne Koller**
Department of Computer Science
Stanford University
Stanford, CA 94305-9010
e-mail: *koller@cs.stanford.edu*

## Abstract

We consider probabilistic inference in general hybrid networks, which include continuous and discrete variables in an arbitrary topology. We reexamine the question of variable discretization in a hybrid network aiming at minimizing the information loss induced by the discretization. We show that a nonuniform partition across all variables as opposed to uniform partition of each variable separately reduces the size of the data structures needed to represent a continuous function. We also provide a simple but efficient procedure for nonuniform partition. To represent a nonuniform discretization in the computer memory, we introduce a new data structure, which we call a Binary Split Partition (BSP) tree. We show that BSP trees can be an exponential factor smaller than the data structures in the standard uniform discretization in multiple dimensions and show how the BSP trees can be used in the standard join tree algorithm. We show that the accuracy of the inference process can be significantly improved by adjusting discretization with evidence. We construct an iterative anytime algorithm that gradually improves the quality of the discretization and the accuracy of the answer on a query. We provide empirical evidence that the algorithm converges.

## 1 INTRODUCTION

A *Bayesian network* is an efficient representation of a joint probability distribution over domain variables. Bayesian networks allow intuitive causal interpretation of dependencies as well as efficient algorithms for probabilistic inference. In particular, we can obtain answers for queries about the probabilities of some events given information about others. Bayesian networks are becoming a popular tool for reasoning under uncertainty and have been used in a number of practical systems.

Although there exists a number of efficient inference algorithms and implementations for probabilistic reasoning in Bayesian networks with discrete variables (for example, [Pearl, 1988, Lauritzen and Spiegelhalter, 1988, Li and D'Ambrosio, 1994, Dechter, 1996]), few algorithms support efficient inference in *hybrid Bayesian networks*, Bayesian networks where continuous and discrete variables are intermixed. Exact probabilistic inference in hybrid networks can be reduced to taking multidimensional integrals in the same way that exact inference in discrete networks can be reduced to computing sums [Li and D'Ambrosio, 1994, Dechter, 1996]. However, computing integrals exactly is possible only for a restricted class of continuous functions.

For example, one of the hybrid Bayesian network classes where exact probabilistic inference is possible are networks with *Conditional Gaussian* (CG) density functions [Lauritzen and Wermuth, 1989, Lauritzen, 1992, Olesen, 1993]. Probabilistic inference in these networks is polynomial in the number of continuous variables. However, the CG limitations on the dependencies between variables obstruct the application of hybrid networks in many domains. In many practical problems we have dependencies substantially different from those encompassed by the CG model. For example, we cannot model a discrete sensor of a continuous variable, say a fire alarm activated by smoke concentration, since continuous variables are not allowed to have discrete descendants in CG hybrid networks.

Given that we want to reason with a more general class of networks and distributions than CG, we need to design approximate methods for inference in hybrid networks. A useful extension of the previous technique is to decompose an arbitrary conditional probability distribution into several CG distributions, and to represent continuous functions as the sums of CG functions [Driver and Morrel, 1995, Alag and Agogino, 1996]. The price one pays is a fast growth of the number of terms in the sums during probabilistic inference. In a join tree, for example, each time a clique potential is multiplied by a message, the number of terms in the resulting sum is the product of the number of terms in each of the factors before the multiplication, thus making the number of terms in the sums grow exponentially with the path length. In addition to this problem, the initial approximation of an arbitrary continuous function by a sum of CGs can also present a *computational challenge*. The number of terms in such a decomposition can be prohibitively large, and probabilistic inference computationally intractable.

The other approach, the one most commonly used in practice, is to discretize all variables in a network. Traditionally, we would discretize each variable separately and represent the conditional probabilities of the nodes in a network and the clique potentials as multidimensional tables. In this approach, the size of a clique potential table is the product of the number of discretized values for the variables participating in the clique. Since the number of variables in a clique can be quite large, we usually cannot afford to discretize the variables as finely as we would like. As a consequence, we incur significant error relative to the exact solution.

In this paper, we propose an alternative approach. Rather than discretizing each variable separately, we discretize a continuous function on its entire multidimensional domain at once. Thus, we can adjust our discretization to the shape of the function, providing a finer partition in places where the function changes rapidly, while leaving relatively "flat" regions at a very rough level of granularity. We show that the nonuniform discretization allows us to provide much greater accuracy using the same number of partitions of the function domain.

To represent one of such nonuniform discretization, we introduce a new data structure, a Binary Split Partition (BSP) tree. A BSP tree represents a recursive binary partition of a function domain and is similar to the quadtrees or octrees used in graphics for representing space objects [Samet and Webber, 1988].[1] In a BSP tree, we restrict the partitions of the multidimensional domains to binary splits by a plane orthogonal to one of the coordinate axes. We show that, for a given number of partitions, BSP trees come very close to the optimal nonuniform discretization of a *multidimensional probability function*. In particular, they are significantly more compact than the traditional representation of continuous functions by multidimensional tables.

Of course, we don't want to discretize the entire joint density function at once. We therefore propose an approach where the density function in each clique in a join tree is discretized using our nonuniform discretization. We show how the traditional join tree algorithm can be adapted to do inference with the BSP trees, and how the probabilistic inference steps can be appropriately interleaved with discretization steps.

Like most other approximation techniques, our approach targets the discretization to do well on the most likely scenarios. As a consequence, it is likely to incur large errors in the case of unlikely evidence. We could refine our discretization so that it remains accurate under all circumstances, but the resulting discretization is likely to be infeasibly large. Rather, we propose an approach that adjusts the discretization to reflect the observed evidence and the needs of the query. An optimal implementation of this process requires that we base our discretization on the posterior probability distributions. Since we do not have access to this posterior, we execute an iterative procedure, in which the accuracy of our predictions and the quality of our discretization increases with each iteration. We show empirically that the results of this procedure converge quickly to the exact results.

## 2    NONUNIFORM DISCRETIZATION

A discretization is conventionally understood as a subdivision of the range of a continuous variable into a set of subranges. If each of the variables is discretized separately, the computational complexity of probabilistic inference grows as the number of discretization subranges, or the number of states per variable, to the power of the induced width of the graph [Dechter, 1996]. Since the induced width can be as large as 20 for practical networks, we want to keep the number of discretization subregions low while preserving most of the information in the discretized function.

We would like to discretize only the regions of a function that contribute to the structure of the joint probability distribution. By providing more detail about the more relevant parts of the space, we can provide a much more accurate picture of the distribution than if we discretize each variable separately. In this section, we generalize the notion of a discretization. We begin by considering the problem of discretizing a single function on a multidimensional domain. In later sections, we will apply these ideas to decomposed probability distributions, such as those found in a Bayesian network.

All treatment in this part is based on the relative entropy or Kullback-Leibler (KL) distance between two probability density functions $f(x)$ and $g(x)$ [Cover and Thomas, 1991]:

$$D(f\|g) = \int_S f(x) \log \frac{f(x)}{g(x)}\, dx \qquad (1)$$

as a metric for the error introduced by the discretization.    There are many justifications for the use of relative entropy as a distance metric between distributions [Cover and Thomas, 1991], including several of useful properties that we will use throughout the paper.

---

[1]Recursive partition of multivariate domains have been also used in multivariate regression and machine learning [Breiman et al., 1984, Moore, 1991].

Our first task is to find the optimal discretization and optimal values for the discretized function. Without loss of generality, we consider discretizing a continuous function defined on a hypercube $\Omega = [0, 1]^n$. To compare the results for different discretizations, we need a formal definition.

**Definition 2.1:** A discretization $\mathcal{D}$ of a hypercube $\Omega = [0, 1]^n$ is a piecewise constant function $i_{\mathcal{D}}(x_1, \ldots, x_n)$ from $\Omega$ to a finite set of integers from 1 to $m$. The function defines mutually exclusive and collectively exhaustive set of subregions $\{\omega_i, i = 1, \ldots, m\}$ in $\Omega$. ∎

If a probability density function $f_{\mathcal{D}}(x_1, \ldots, x_n)$ is constant in each of the subregions $\omega_i$, we call it a discretized function on the discretization $\mathcal{D}$. The following theorem proves that the optimal value for the discretized function $f_{\mathcal{D}}(x_1, \ldots, x_n)$ is the mean of the function $f(x_1, \ldots, x_n)$ in each of the subregions $\omega_i$.

**Theorem 2.1:** *Given a discretization $\mathcal{D}$ of a region $\Omega$, the minimum KL distance between a probability density function $f(x_1, \ldots, x_n)$ and a discretized function $f_{\mathcal{D}}(x_1, \ldots, x_n)$ is achieved by the discretized function $\tilde{f}_{\mathcal{D}}(x_1, \ldots, x_n)$ which is piecewise constant and in each of the subregions $\omega_i$ is equal to the mean of the function $f(x_1, \ldots, x_n)$ in the corresponding subregion $\omega_i$. The KL distance to any other piecewise constant on $\mathcal{D}$ function $f_{\mathcal{D}}(x_1, \ldots, x_n)$ which is also piecewise constant and in each of the subregions $\omega_i$ is given by the sum of KL distances from $f_{\mathcal{D}}(x_1, \ldots, x_n)$ to $\tilde{f}_{\mathcal{D}}(x_1, \ldots, x_n)$ and from $\tilde{f}_{\mathcal{D}}(x_1, \ldots, x_n)$ to $f(x_1, \ldots, x_n)$.*

This is a consequence of well-known decomposition properties of relative entropy distance [Cover and Thomas, 1991]. Due to the space restrictions we do not show the proof of the theorem.

Now that we have a procedure to assign values to a discretized function given the discretization, we need to find a discretization that minimizes the relative entropy error. Below, we provide a simple divide and conquer technique for finding a discretization which is very close to optimal.

To find an optimal discretization of a hypercube $\Omega$ into $m$ subregions we should search through all possible partitions of $\Omega$ into $m$ subregions for a partition that gives us the minimal KL distance between the original function $f(x_1, \ldots, x_n)$ and its discretized function $f_{\mathcal{D}}(x_1, \ldots, x_n)$ with optimally assigned values in the subregions. This procedure is computationally intensive even in one dimension and becomes computationally intractable in multiple dimensions. Instead, we propose a simple recursive technique and show that it generates results very close to optimal.

We borrow the idea of a recursive space decomposition from graphics [Samet and Webber, 1988], where quadtree and octree decomposition of two- and three-dimensional spatial respectively are often used for a compressed representation of space objects. Both of these data structures have proved to be very efficient computationally.

A Binary Split Partition (BSP) tree is a recursive data structure that represents a hierarchical binary decomposition of a multidimensional function. Each node in the tree represents a subregion of the function domain and can have two children. Each of the children represents half of the parent's space. The splitting continues from the root of the tree, representing the whole function domain, to the leaves, that carry information about the function in a particular subregion.

We restrict the splits to subdivisions of the space into two halves defined by a plane orthogonal to one of the axes. For example, one of the possible BSP trees for a function of two variables $x$ and $y$ is shown in Fig. 1. On the first level, we split the function domain by a line orthogonal to $y$. On the second level, we leave the left node as a leaf representing the lower half of the $xy$ plane. We split the right one, representing the upper half of the $xy$ plane, by a line orthogonal to $x$. Each of the children on the third level is split even further.

As a result, the function represented by the BSP tree in Fig. 1 is constant in the lower half of the $xy$ plane. The discretization has higher granularity in the upper half of the $xy$ plane, where we continue splitting. The BSP tree in Fig. 1 might be useful for representing a function with some structure in the upper half of the $xy$ plane.

To discretize a function, we need heuristic that tells us how to build up a BSP tree: which leaf to split next and in which direction. Since computing the exact contribution to the relative entropy error is computationally expensive for a general function, we use a bound on the KL distance between the function $f$ and its discretization $f_{\mathcal{D}}$ based on the function mean $\bar{f}$, the function maximum $f_{max}$, and the function minimum $f_{min}$ in the given subregion $\omega_i$:

$$\int_{\omega_i} f \log \frac{f}{\bar{f}} d\Omega \leq \Big[ \frac{f_{max} - \bar{f}}{f_{max} - f_{min}} f_{min} \log \frac{f_{min}}{\bar{f}} + \frac{\bar{f} - f_{min}}{f_{max} - f_{min}} f_{max} \log \frac{f_{max}}{\bar{f}} \Big] |\omega_i|,$$

where $|\omega_i|$ denotes the volume of a discretization subregion $\omega_i$. The parameters $\bar{f}$, $f_{max}$, $f_{min}$ are estimated by randomly sampling $f$ at several points.

During the discretization process, all leaves are kept in a priority queue. The estimates of the relative entropy error are used to take the leaves out of the queue. A leaf with the largest error estimate is then split first, and the two resulting leaves are put back into the queue. To control the accuracy of our discretization, we also maintain the sum of all estimates for all the leaves in the queue. We stop the discretization process when either the estimate of the error becomes smaller than some fixed parameter $\delta$ or the number of leaves exceeds some fixed number $N$.

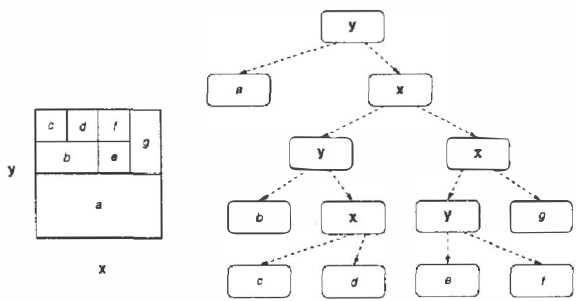

Figure 1: An example of a two dimensional hierarchical space decomposition. Internal nodes of the tree store the axis of a split. Leaves of the tree store the average of the continuous density function over the subregion represented by the leaf.

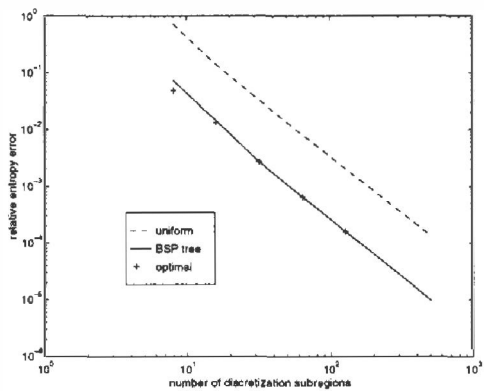

Figure 3: The relative entropy error of discretization as a function of the number of discretization subregions for equidistant (dashed line), BSP tree (solid line), and optimal (crosses) discretization. The optimal discretization was found by the gradient descent method. The error of the BSP tree and optimal discretizations are almost identical for large number of discretization intervals. The error of uniform discretization is about a factor of ten larger.

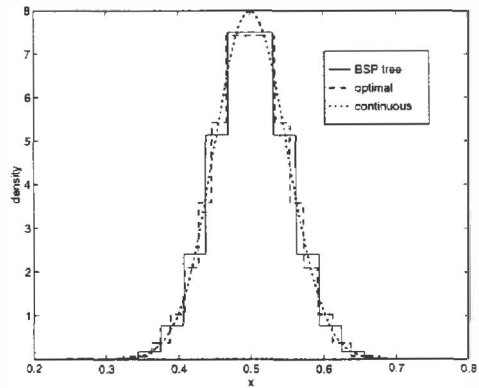

Figure 2: BSP tree (solid line) and optimal (dashed line) discretization of a normal distribution $N(x; 0.5, 0.0025)$ (dotted line). The number of discretization intervals is 16 in both cases. The optimal discretization was found by the gradient descent method.

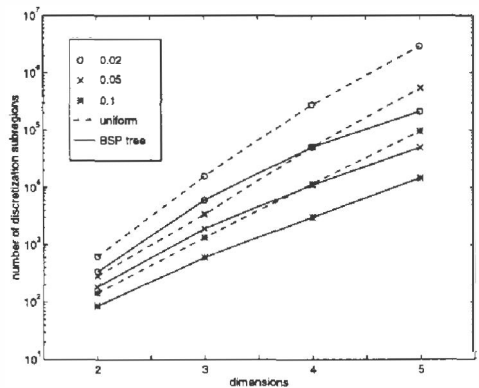

Figure 4: Number of discretization subregions as a function of the number of dimensions for BSP tree (solid line) and uniform (dashed line) discretization. The discretization was performed to approximate a multivariate normal distribution proportional to $N(\sum_1^{n-1} x_i/(n-1) - x_n; 0, 0.0025)$ with relative entropy error 0.02, 0.05, and 0.1. For a large number of dimensions, the BSP discretization performs much better (notice the logarithmic scale for the number of discretization subregions).

Finding the direction of the optimal split presents a challenging problem. In an ideal situation, we would estimate the decrease in the relative entropy distance due to all possible splits and choose the optimal one. However, to do this exactly, we would need to estimate multidimensional integrals. Instead, we sample several points around the center of the subregions $\omega_i$ and pick the direction in which the function changes most, i.e., the coordinate axis along which the ratio $f_{max}/f_{min}$ is the largest around the center of the subregion $\omega_i$.

The result of the one-dimensional discretization of a normal distribution $N(x; \mu, \sigma^2) = \frac{1}{\sqrt{2\pi}\sigma} \exp(-(x-\mu)^2/2\sigma^2)$ with $\mu = 0.5$ and $\sigma = 0.05$ is shown in Fig. 2. The algo-rithm correctly chooses to discretize the regions that have higher derivative and/or are high in probability density. In fact, the discretization obtained with BSP tree is very close to the optimal discretization obtained by the gradient descent method. This is confirmed by Fig. 3, which shows the relative entropy error as a function of the number of discretization subregions. The error of the BSP tree discretization and the optimal discretization is virtually the same for the number of discretization intervals larger than 16. On the other hand, an equidistant discretization requires about a factor of 5 more discretization intervals to reach the same accuracy.

BSP trees are even more promising for representing multidimensional density functions. If a function has sharp ridges, the savings are exponential in the number of

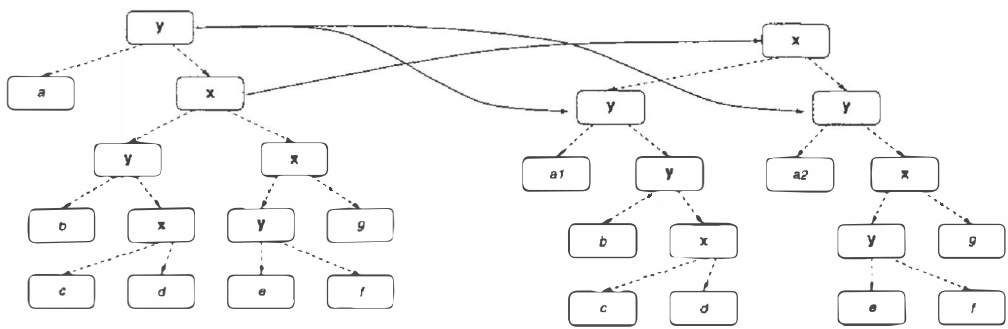

Figure 5: Adjusting the structure of the BSP tree in Fig. 1 to another BSP tree that has a root split on variable $x$.

dimensions—we save a constant factor along each of the dimensions. The results of discretizing a multivariate normal distribution proportional to $N(\sum_1^{n-1} x_i/(n-1) - x_n; 0, 0.0025)$ with different relative entropy error are shown in Fig. 4. Given the accuracy, the number of discretization subregions grows much slower for a BSP discretization than for a standard uniform equidistant discretization of each of the variables separately. We save about a factor of 10 in 5 dimensions.

## 3    OPERATIONS ON BSP TREES

In this section, we briefly consider summation, multiplication, and integration of functions represented by BSP trees. We will show in the next section how the BSP trees can be used in the standard join tree algorithm for probabilistic inference.

We assume that the BSP trees are encoded as a tree structure that uses pointers. Other implementations, that might be more computationally efficient, are possible, but are out of the scope of this paper. We refer the reader to [Samet and Webber, 1988] for a comprehensive review on this subject.

Let us start with the summation of two BSP trees. We will use summation later in our integration algorithm. If the structure of the trees is aligned, i.e., they have exactly the same splits on the same levels, the summation is reduced to a tree traversal. The values at the leaves are summed during the traversal. Tree traversal takes linear time and the computational complexity for the aligned trees is $O(N)$, where $N$ is the number of leaves in a tree.

If the trees are not aligned, we need to adjust the structure of one of them by inserting additional nodes. For example, if we want to sum the tree in Fig. 1 with another tree that has a root split on variable $x$, we adjust the structure of the first tree as shown in Fig. 5 by moving the $x$ split in the right branch up and by making two additional leaves in the corresponding branches. Complete alignment of the trees takes $O(N_1 \times N_2)$ operations in the worst case, where $N_1$ and $N_2$ are the number of leaves in the first and the second trees respectively.

```
1:  input: two nodes of a tree representing the same subregion ω,
2:  output: a node of a tree representing the result of summation in
    the subregion ω,
3:  if both nodes are leaves then
4:     return a leaf with the sum of the values stored in the nodes
5:  else if the second node is a leaf then
6:     add the constant from the second node to all leaves of the first
       subtree
7:     return the first node
8:  else
9:     if split on different variables then
10:        adjust the structure of the first tree
11:    end if
12:    add left subtrees of both operands
13:    add right subtrees of both operands
14:    return a node with the result of the previous two operations
       as its children
15: end if
```

Figure 6: Algorithm for the summation of two BSP trees.

The algorithm for summing two nodes of a BSP tree is shown in Fig. 6. Summation takes $O(N_1 + N_2)$ operations in the best and $O(N_1 \times N_2)$ operations in the worst case. Intuitively, if all the splits in both operands are on the same variable, the computational complexity is linear. If the trees are completely misaligned, for example if all the splits in the first tree are on variable $x$ and in the second tree are on variable $y$, then the computational complexity is quadratic.

```
1:  input: two nodes of a tree representing the same subregion ω,
2:  output: a node of a tree representing the result of multiplication
    in the subregion ω,
3:  if both nodes are leaves then
4:     return a leaf with the product of the values stored in the nodes
5:  else if the second node is a leaf then
6:     multiply all leaves in the first subtree by a constant from the
       second node
7:     return the first node
8:  else
9:     if split on different variables then
10:        adjust the structure of the first tree
11:    end if
12:    multiply left subtrees of both operands
13:    multiply right subtrees of both operands
14:    return a node with the result of the previous two operations
       as its children
15: end if
```

Figure 7: Algorithm for multiplying two BSP trees.

The algorithm for multiplying two nodes of a BSP tree is shown in Fig. 7. As in the summation algorithm, we have to align the trees when they have different structure. Analogously, multiplication takes $O(N_1 + N_2)$ operations in the best and $O(N_1 \times N_2)$ operations in the worst case when the trees are completely misaligned.

```
1:  input: a node of a tree and the i-th variable to be integrated over
2:  output: a node of the tree representing the result of the integration
3:  if the node is a leaf then
4:      return the node itself
5:  else if split on i-th variable then
6:      integrate left subtree
7:      integrate right subtree
8:      return the sum of the left and the right subtrees divided by two
9:  else
10:     return the node itself
11: end if
```

Figure 8: Algorithm for integrating a function represented as a BSP tree over a variable.

Finally, we consider the integration of a function represented as a BSP tree over some variable. This is the operation that, in the discrete case, corresponds to variable elimination by summation. Since a leaf represents a constant value of a function over a region, integration of a leaf is reduced to multiplication of the value stored in the leaf by the remaining multidimensional volume. It is possible to compute this volume during the process of tree traversal, since the volume of a subregion represented by a child is always half the size of the subregion represented by its parent. The integration algorithm is presented in Fig. 8. Integration also takes $O(N_1 + N_2)$ operations in the best and $O(N_1 \times N_2)$ operations in the worst case.

Many other algorithms, for such tasks as computing the expected value of a function, the cross entropy, or the differential entropy, can be expressed as a simple traversal of the tree, thus taking linear time with respect to the size of the tree.

## 4  BASIC PROBABILISTIC INFERENCE ALGORITHM

We now show how we can integrate our nonuniform discretization using BSP trees into standard Bayesian network inference algorithms. Our approach will be based on the *optimal factoring approach* to probabilistic inference [Li and D'Ambrosio, 1994].

We illustrate our algorithm using an example. Let us assume that we can observe a one-dimensional robot on the interval $0 \leq x \leq 1$. Although we do not know the robot's position exactly, we know the readings of a number of sensors and know that the robot can walk randomly between the observations. The Bayesian network corresponding to this situation is shown in Fig. 9.

If the robot's position $x$, the first and the second sensor reading is $o$ with probability $p(o|x) = N(o - x; 0, 0.01)$. Thus, the first two observations are noisy observations of

the robot coordinate. The third observation is a discrete noisy observation of the robot in the left halfspace $x < 0.5$. If the robot position is $x$, the sensor is likely to give a reading of *true* with probability $\left(1 + \exp(40(x - 0.5))\right)^{-1}$.

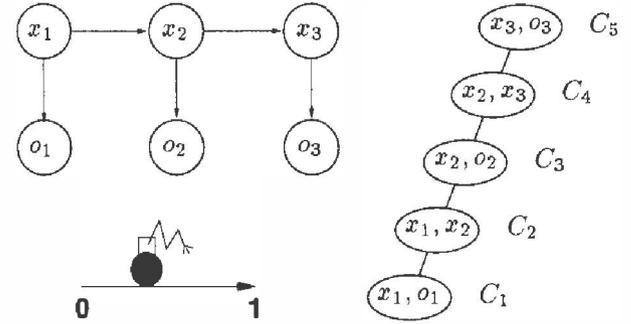

Figure 9: A simple hybrid Bayesian network used as an example and its join tree.

The dependence of the robot coordinate at the next observation on the robot position at the current observation is given by the conditional probabilities $p(x_3|x_2) = N(x_3 - x_2; 0, 0.01)$ and $p(x_2|x_1) = N(x_2 - x_1; 0, 0.01)$. Our prior beliefs about the robot position are uniform: $p(x_1) = 1$. We would like to know the probability distribution of the robot coordinate after three observations.

The join tree corresponding to this network is shown in Fig. 9. Probabilistic inference with continuous variables can be analyzed in the same *optimal factoring approach* used in[Li and D'Ambrosio, 1994]. The posterior probability of the robot coordinate after observations $o_1$, $o_2$, and $o_3$ is:

$$
\begin{aligned}
p(x_3|o_1, o_2, o_3) &\equiv p(x3, o1, o2, o3)/p(o1, o2, o3) \\
&\sim \int_0^1 \int_0^1 p(o_3|x_3)p(x_3|x_2)p(o_2|x_2)p(x_2|x_1) \\
&\qquad\qquad p(o_1|x_1)p(x_1)\,dx_1\,dx_2 \\
&= p(o_3|x_3) \int_0^1 p(x_3|x_2)p(o_2|x_2) \\
&\qquad \left( \int_0^1 p(x_2|x_1)p(o_1|x_1)p(x_1)\,dx_1 \right) dx_2.
\end{aligned}
\tag{2}
$$

Computing integrals in the above decomposition corresponds to computing messages in the standard join tree algorithm. Partial sums or integrated conditional probabilities represent messages passed from one clique to another.

This network was chosen so as to allow us to compare the performance of our algorithm to the true answers. Therefore, the network is exactly solvable up to a normalizing factor given the observations $o_1$, $o_2$, $o_3 = true$:

$$
p(x_3) \sim \frac{N(x_3; (o_1 + 2o_2)/3, 1/60)}{1 + \exp(40(x_3 - 0.5))};
\tag{3}
$$

where the answer was obtained by integrating the joint probability over $x_1$ and $x_2$.

The reformulated join tree algorithm is shown in Fig. 10. For our network, we would first multiply the continuous conditional probabilities $p(o_1|x_1)$ and $p(x_1)$ in clique $C_1$, discretize it using our BSP tree construction algorithm, and pass a discretized message to clique $C_2$. The clique $C_2$ will multiply the message by $p(x_2|x_1)$, discretize it, then integrate over $x_1$, and pass it to the next clique $C_3$. The basic operations used in the inference process reduce to the operations described in Section 3. The process continues until the last clique, $C_5$, receives the discretized message, multiplies it by the continuous function $p(o_3|x_3)$, and discretizes it. Note that the function at a clique is only discretized after it received the messages from its subtrees, allowing its choice of discretization to be much more informed.

```
1: build a join tree through moralization and triangulation of the
   Bayesian network graph
2: assign continuous functions to cliques that completely contain all
   their arguments
3: find a clique that contains the query node and make it the root of
   the tree
4: for each clique starting from the leaves and up to the root do
5:     multiply all messages from descendants
6:     multiply the previous result by the assigned functions
7:     discretize the previous result with some fixed precision δ
8:     form a message up by integrating over variables that are not in
       the parent clique
9: end for
```

Figure 10: A reformulated join tree algorithm.

In many cases, this process does very well. For example, as we see in Fig. 11, if our first two observations are the same—$o_1 = o_2 = 0.2$—the results of the inference are very close to the exact solution, computed analytically in (3). However, as our observations become more and more unlikely, the accuracy of the results of this basic inference algorithm begins to deteriorate. If, for example, the second observation is changed to $o_2 = 0.65$, the results of the inference contain only a single bump (see Fig. 11) and are very different from the exact answer. In the next section we describe how to improve the performance of this algorithm in the case of unlikely evidence.

## 5    DYNAMIC DISCRETIZATION

To understand why the results of probabilistic inference depend on the probability of evidence $p(e)$, let us consider the decomposition of the KL distance between the true joint probability distribution given by the product of all continuous functions $p(x, e) = \prod_i p(x_i | Pa(x_i))$ and the discretized joint probability distribution $f_{\mathcal{D}}(x, e)$:

$$D(p(x, e) \| f_{\mathcal{D}}(x, e))$$
$$= D(p(e) \| f_{\mathcal{D}}(e)) + D(p(x|e) \| f_{\mathcal{D}}(x|e)),$$

where $D(p(x|e) \| f_{\mathcal{D}}(x|e))$ is the conditional relative entropy or conditional Kullback-Leibler distance:

$$D(p(x|e) \| f_{\mathcal{D}}(x|e))$$
$$= \int p(e) \int p(x|e) \log p(x|e) / f_{\mathcal{D}}(x|e) \, dx \, de.$$

The second integral essentially represents a KL distance between the desired answer to a query $p(x|e)$ and the answer obtained with our discretized network $f_{\mathcal{D}}(x|e)$. However, this relative entropy integral is multiplied by $p(e)$. Thus, even if we bound the relative entropy error of the discretization for the whole network by $\epsilon$, the bound on the error of our answer is $\epsilon/p(e)$.

Rather than minimizing the KL distance of the full joint probability, as we do by discretizing partial sums, we need to minimize the KL distance:

$$\int p(q|e) \log p(q|e) / f_{\mathcal{D}}(q|e) \, dq$$

of the query node $q$ conditioned on the evidence $e$. Given the evidence, we might want to change the discretization and rediscretize the regions that become probable given the evidence. Thus, we propose a more general metric that will reflect our preferences to discretize some regions more finely than the others.

**Definition 5.1:** The weighted relative entropy or Weighted Kullback-Leibler (WKL) distance $W(f(x) \| g(x); w(x))$ between the density functions $f(x)$ and $g(x)$ with a strictly positive weight $w(x)$ is defined by:

$$W(f(x) \| g(x); w(x)) = \int_S w(x) f(x) \log \frac{f(x)}{g(x)} \, dx \quad (4)$$

where we assume that the integral exists. ∎

The WKL distance reduces to the KL distance if the weight function is constant. Although in general the WKL distance is not even sign definite, the following inequalities hold if the weight $w(x)$ and the function $f(x)$ are discretized using the same discretization $\mathcal{D}$:

$$D(f(x) \| f_{\mathcal{D}}(x)) \min_x w_{\mathcal{D}}(x)$$
$$\leq W(f(x) \| f_{\mathcal{D}}(x); w_{\mathcal{D}}(x))$$
$$\leq D(f(x) \| f_{\mathcal{D}}(x)) \max_x w_{\mathcal{D}}(x).$$

The proof follows from considering the contributions from each of the subregions $\omega_i$. These inequalities follow from bounding each contribution by maximum/minimum weight times the corresponding contribution to the relative entropy.

### 5.1    WEIGHT ASSIGNMENT

Now, consider our initial discretization procedure. At each clique $C_i$, we discretize some function which combines the incoming messages with the clique's own assigned functions. The problem is that this function only takes into consideration the evidence which is "down the tree." If the other evidence results in a very different posterior distribution, our discretization procedure may be placing the emphasis on the wrong part of the space.

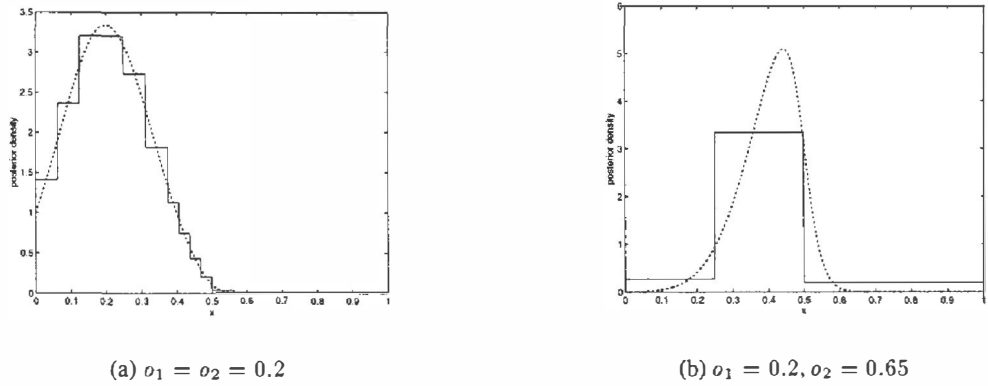

(a) $o_1 = o_2 = 0.2$                    (b) $o_1 = 0.2, o_2 = 0.65$

Figure 11: Posterior probability $p(x_3)$ for a network shown in Fig. 9 for similar (a) and contradictory (b) evidence. The results of the inference are shown by a solid line, the exact result is shown by a dotted line. Evidence $o_3$ is true in both cases.

This is not a problem at the root clique, since the root clique gets messages which already combine the information at all the other cliques. When we combine it with the clique's own assigned functions, the result is the best approximation we have for the clique's correct posterior density function. Therefore, a constant weight at the root clique is the right one. An intermediate clique, on the other hand, receives messages from all but the parent clique, the one that connects it with the root. Had it had this message, the product of all the messages and the assigned clique functions would have been the best guess about the posterior distribution. Again, no weights would have been necessary. Since the clique lacks this additional message from its parent, the function it uses for the discretization is an incomplete estimate of the posterior. In order to get this function as close as possible to the posterior, we must choose a weight $w$ that best approximates this "missing" message from the parent clique.

To make this intuition precise, let us track the errors in the network more rigorously, starting from the root clique containing the query node, and prove that the best weight for a clique is essentially the message that it should have gotten from its parent. Formally, our goal is to assign weights $w$ to the various join tree cliques so that minimizing the WKL distance between the correct and discretized function within each clique will minimize our error for the probability of the query node $q$ given evidence $e$. We assign weights by going *down* the tree, in the direction opposite to the initial message propagation (as in (2)). We assume that the weight for a parent clique are already known, and then pick a weight for the child clique to minimize the resulting WKL distance for the parent.

We begin with the root clique. Our goal at this clique is to minimize the KL distance of the query given the evidence, the weight at that clique should be uniformly 1. Now, let us look at how the weights for a child clique can be derived given the weights for a parent. Consider two cliques $C_1 = \{x, y\}$ and $C_2 = \{y, z\}$, where $C_1$ is the parent and $C_2$ the child. Let $w(x, y)$ be the weight for clique $C_1$. We want to find the best weight on the message $s(y)$ coming from the clique $C_2$ that minimizes the relative entropy error of the true potential $f(x, y) = r(x, y)s(y)$, where $r(x, y)$ stands for the product of all assigned to the clique functions and $s(y)$ for the message coming from $C_2$, relative to the discretized potential $f_{\mathcal{D}}(x, y) = r_{\mathcal{D}}(x, y)s_{\mathcal{D}}(y)$. We decompose the WKL distance between $f$ and $f_{\mathcal{D}}$ into a sum of WKL distances:

$$
\begin{aligned}
W & (f(x, y) \| f_{\mathcal{D}}(x, y); w(x, y)) \\
&= \int w(x, y) f(x, y) \log \frac{f(x, y)}{f_{\mathcal{D}}(x, y)} \, dx \, dy \\
&= \int w(x, y) r(x, y) s(y) \log \frac{r(x, y) s(y)}{r_{\mathcal{D}}(x, y) s_{\mathcal{D}}(y)} \, dx \, dy \\
&= \int w(x, y) r(x, y) s(y) \log \frac{r(x, y)}{r_{\mathcal{D}}(x, y)} \, dx \, dy \\
&\quad + \int w(x, y) r(x, y) s(y) \log \frac{s(y)}{s_{\mathcal{D}}(y)} \, dx \, dy \\
&= W \left( r(x, y) \| r_{\mathcal{D}}(x, y); w(x, y) s(y) \right) \\
&\quad + W \left( s(y) \| s_{\mathcal{D}}(y); \int w(x, y) r(x, y) \, dx \right).
\end{aligned}
$$

In our discretization procedure (see Fig. 10), we first get the discretized message $s_{\mathcal{D}}(y)$, and then discretize the product $r(x, y) s_{\mathcal{D}}(y)$. Since $r_{\mathcal{D}}(x, y) = f_{\mathcal{D}}(x, y)/s_{\mathcal{D}}(x, y)$, the clique's potential discretization procedure is responsible for the first term in the sum and is controlled by the pre-cision parameter $\delta$. The minimization of the second term:

$$W\left(s(y)\|s_{\mathcal{D}}(y); \int w(x,y)r(x,y)\,dx\right)$$
$$= W\left(s(y)\|s_{\mathcal{D}}(y); \frac{\int w(x,y)f(x,y)\,dx}{s(y)}\right) \quad (5)$$

is implicitly done when we discretize $C_2$'s potential and is independent of the discretization of $C_1$'s potentials. Thus, by reducing the WKL distance (5) we reduce the WKL distance of the true potential to the discretized potential of the clique $C_1$ which will be used to propagate message further to the root. This result can be easily extended to several child cliques; in this case $r(x, y)$ is the product of all assigned to the clique functions and messages from all the other children.

Similarly, considering the clique $C_2$ which has to integrate its potentials to form the message $s(y) = \int f(y, z)\,dz$:

$$W\left(f(y,z)\|f_{\mathcal{D}}(y,z); w(y,z)\right)$$
$$= \int w(y,z)f(y,z)\log\frac{f(y,z)}{f_{\mathcal{D}}(y,z)}\,dy\,dz$$
$$= \int w(y,z)f(y,z)\log\frac{s(y)}{s_{\mathcal{D}}(y)}\,dy\,dz$$
$$+ \int w(y,z)f(y,z)\log\frac{f(y,z)/s(y)}{f_{\mathcal{D}}(y,z)/s_{\mathcal{D}}(y)}\,dy\,dz,$$

we can derive that by reducing the WKL distance $W\left(f(y,z)\|f_{\mathcal{D}}(y,z); w(y,z)\right)$ we reduce the WKL distance:

$$W\left(s(y)\|s_{\mathcal{D}}(y); \frac{\int w(y,z)f(y,z)\,dz}{s(y)}\right), \quad (6)$$

of the true message $s(y)$ to the discretized one $s_{\mathcal{D}}(y)$ with the weight $\int w(y,z)f(y,z)/s(y)\,dz$.

## 5.2   WEIGHT PROPAGATION

Comparing equations (5) and (6), we conclude that the weights of the neighboring cliques should satisfy the following condition:

$$\frac{\int w(x,y)f(x,y)\,dx}{s(y)} = \frac{\int w(y,z)f(y,z)\,dz}{s(y)}. \quad (7)$$

This equation essentially says that the products of weights and clique potentials of the two neighboring cliques have to be calibrated. Given that the weight and the clique potential of clique $C_1$ is fixed, equation (7) tells us to choose $w(y, z)$ to guarantee calibration of clique $C_2$ to the clique $C_1$. But this is exactly the process for propagating the posterior evidence back to the leaves in the join tree algorithm, except that we update the weights so that the product of the clique's weight and potential is calibrated, not the clique potential itself.

A very similar result can be achieved by considering the decomposition of the KL distance of the true joint probability $f(x_1, \ldots, x_n)$ of a network, which is the product of all clique potentials $f^{C_i}(X_i)$, to the discretized joint probability $f_{\mathcal{D}}(x_1, \ldots, x_n)$, which is the product of all discretized potentials $f_{\mathcal{D}}^{C_i}(X_i)$:

$$D(f(x_1, \ldots, x_n)\|f_{\mathcal{D}}(x_1, \ldots, x_n))$$
$$= \sum_i \int_\Omega f(x_1, \ldots, x_n)\log\frac{f^{C_i}(X_i)}{f_{\mathcal{D}}^{C_i}(X_i)}\,d\Omega, \quad (8)$$

where we denoted the set of variables in the clique $C_i$ as $X_i$. Equation (8) is a consequence of the decomposition properties of the KL distance (1) and says that the weight for the clique $C_i$ should be $\int_{x_k \notin X_i} f(x_1, \ldots, x_n)\,d\Omega/f^{C_i}(X_i)$. Calibration (7) is different because we discretize the product of the clique potential and messages from the child cliques, not the clique potential itself.

We note that we made several assumptions during our derivation of (7). For instance, we approximated $f_{\mathcal{D}}(x, y)$, which is a discretized product $f(x, y) = r(x, y)s(y)$, by a product $r_{\mathcal{D}}(x, y)s_{\mathcal{D}}(y)$ of discretized functions. But a discretized function assigns each of its discrete values a value which is the average of the function in the corresponding region; and it is not generally true that the average of a product $r(x, y)s(y)$ is the product of the averages. More careful analysis shows that for continuous functions the error made by this approximation contributes $o(1/N)$, where $N$ is the average number of splits along a variable, compared to the magnitude of the WKL distance itself. Thus, this error becomes negligible when the BSP trees grow larger.

Finally, we observe that (7) extends to a more general problem than minimizing the standard KL distance of the answer to a query. Consider, for example, a user who is interested in minimizing the WKL distance with weights determined by sensitivity analysis or utility considerations. In this case, the above propagation rule applies naturally, and can be used to provide a good discretization with respect to this particular WKL distance as our error metric.

## 5.3   ITERATIVE ALGORITHM

To apply the idea of discretizing each clique based on the appropriate WKL distance, we need to know the weights. Before any propagation takes place, we clearly do not have this information. However, in order to do any propagation, we need some initial discretization. We avoid the circularity in this definition by using an iterative algorithm. We start out by assigning constant weights (one) to all cliques, and doing an initial round of propagation. In that round, we propagate partial sums up the tree and weights down the tree. When the initial propagation is finished, the cliques have weights, so we can do another round of propagating messages up the tree.

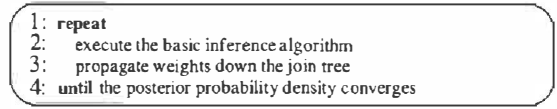

```
1:  repeat
2:      execute the basic inference algorithm
3:      propagate weights down the join tree
4:  until the posterior probability density converges
```

Figure 12: An iterative join tree algorithm.

We repeat this procedure iteratively[2] to get better and better estimates of the weights and more precise answers to our query until the posterior probability of the query node converges (see Fig. 12). The weights are stored in the same BSP tree as the discretized clique potential, thus ensuring the same discretization for the weights and potentials and the non-negativity of the WKL distances.

A BSP tree stores discretization $\mathcal{D}$ at each clique. Given that the trees in our algorithm are not pruned, the discretization can become only finer and we can not effectively change algorithm focus even if the contribution of some discretization subregions to the total WKL distance becomes very small. To avoid the uncontrolled growth of the BSP trees, we prune them on each iteration by removing the leaves that contribute to the total relative entropy error less that an average leave in the tree. The error in the leaves is estimated by Monte Carlo integration during clique discretization.

After pruning, the clique potentials are rediscretized again on the next propagation iteration. A removed leave can reappear again as a result of the clique rediscretization. However, it is less likely to appear if the assigned to the corresponding subregion weight is small. On the other hand, the subregions with the large weights are more likely to be the first in the discretization queue and to be rediscretized more finely. Let us look how this scheme works in practice.

# 6   EXPERIMENTS

All our results are based on the simple problem described in Fig. 9, which has exact solution (3). Errors were evaluated by numerical integration.

## 6.1   REDISCRETIZATION

First, we tested how the discretization adapts to the weights and posterior distributions. We used an unlikely evidence set $o_1 = 0.2$, $o_2 = 0.8$, and $o_3 = true$ (the probability of evidence is only $10^{-3}$) and the precision parameter $\delta = 0.02$. On the first round of propagation, we could not resolve any of the structure; the BSP tree contained only one leaf, reflecting a very poor initial discretization which was done with uniform weights.

Already on the second iteration, after only one phase of weight propagation, the cliques had a very good estimate of the posterior distribution and therefore the weights. The BSP tree on the second iteration had 11 leaves, and the posterior probability distribution differed from the true probability distribution by a KL distance of $0.03$. The BSP tree after the third round of propagation had 18 leaves, and the posterior probability distribution differed from the true probability distribution by a KL distance of $0.001$ (see Fig. 13).

The prior probability discretizations before and after the first weight update are shown in Fig. 14. While at the initial propagation the $N(x_1; 0.2, 0.01)$ Gaussian corresponding to the product $p(o_1|x_1)p(x_1)$ is discretized with the weight one, the weight is nonuniform for the second round of propagation as shown in Fig. 14(b). The new discretization takes into account much larger weights on the right slope and rediscretizes it more finely.

## 6.2   CONVERGENCE

The algorithm converges by the second iteration in most cases. Figure 15 shows the relative entropy error as a function of the iteration number for several precision parameters $\delta$. The relative entropy error dropped very abruptly after the first iteration and experienced small oscillations around the final answer after that.

Pruning allows to compare the efficiency of our approach to the uniform discretization. Since we can effectively change discretization focus with evidence, we get a smaller error with our approach than with a uniform approach for the equivalent number of discretization subregions per clique. Figure 16 shows the relative entropy error of a dynamically discretized network compared to a uniformly discretized network given the same number of partitions per clique (so that if we have had $N$ partitions per variable in a uniform discretization, we would have $N^2$ partitions per two-dimensional clique).

For the nonuniform discretization, we get a factor of 4 better precision as compared to the case of uniform discretization with the same number of discretization subregions. In practice, the savings can be much bigger since our algorithm can focus the discretization effort on the cliques that have potentials most important for a particular evidence, not potentials of all cliques at the same time. However, comparison with the uniform discretization is not so straightforward in this case. Although in our simple experiments we did not get any computation time advantage over a uniform discretization—table representation of conditional probabilities is very efficient computationally—we had evidence in Fig. 4 that the nonuniform dynamic dis-

---

[2]A similar idea of executing inference on a simplified network and then refining the approximation based on the results was also used by [Wellman and Liu, 1994] for the related problem of state-space abstraction.

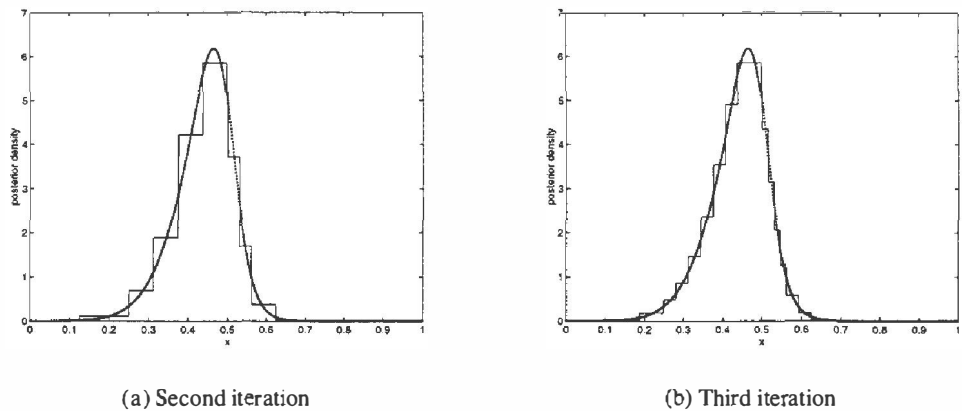

(a) Second iteration                    (b) Third iteration

Figure 13: Posterior probability $p(x_3)$ for a network shown in Fig. 9 with dynamic discretization for two successive iterations. The result of the inference is shown by a solid line, the exact result is shown by a dotted line. Evidence is $o_1 = 0.2$, $o_2 = 0.8$, and $o_3 = true$.

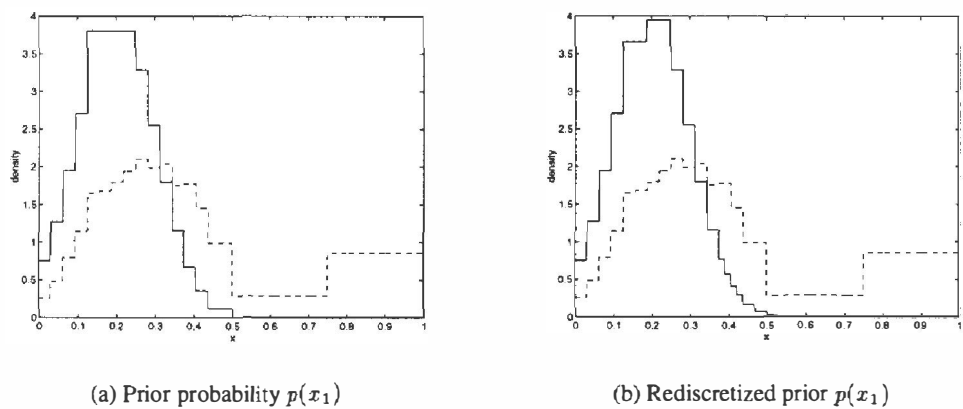

(a) Prior probability $p(x_1)$            (b) Rediscretized prior $p(x_1)$

Figure 14: Original and rediscretized prior probabilities $p(x_1)$. The dashed line shows the estimate of posterior that the clique has after the first weight propagation. Notice the change in the granularity of the discretization. Evidence is $o_1 = 0.2$, $o_2 = 0.8$, and $o_3 = true$.

cretization should be more efficient for more complex domains requiring precision and thus a very fine discretization of multidimensional domains.

## 7   CONCLUSIONS

In this paper, we provide an effective algorithm for probabilistic inference in hybrid networks with an arbitrary topology and arbitrary functional dependence between variables based on nonuniform dynamic discretization. While our approach is based on discretizing the function, it is derived from the key insight that, within the domain of the function, not all of the regions should be accorded equal importance. We suggest the idea of nonuniform discretization, which discretizes multidimensional domains as a whole, rather than discretizing each variable separately. Nonuniform discretization has been successfully used in multivariate regression and machine learning. We provide results for the discretization of probability distributions for probabilistic inference which show that nonuniform discretization can be substantially more compact than the traditional uniform discretization.

Any fixed discretization, however, cannot account well for all possible configurations of evidence. Therefore, we are likely to get large errors for unlikely evidence. We develop a new metric based on the relative entropy that allows to emphasize discretization of some regions as opposed to others and show how to use it in a self-adjusting anytime rediscretization algorithm. This algorithm constantly updates the discretization in accordance with the evidence. It can be run to provide answers of any desired accuracy. Our preliminary empirical results suggest that convergence to the right solution is very rapid in practice.

Given the recent emphasis on building and using hybrid systems, we believe that probabilistic inference algorithms for the corresponding models will become more and more necessary and precision more and more important. As such systems are typically fairly complex, and involve tightly coupled discrete and continuous elements, exact algorithms are unlikely to be available. Our algorithm deals effectively with arbitrary hybrid systems, so that we can hope that it will be applicable to many of these applications. In particular, we believe that the ability of our algorithm to adjust itself to the evidence it sees will prove very useful in applications, e.g., real-time monitoring of hybrid systems, that require fast and efficient change of focus.

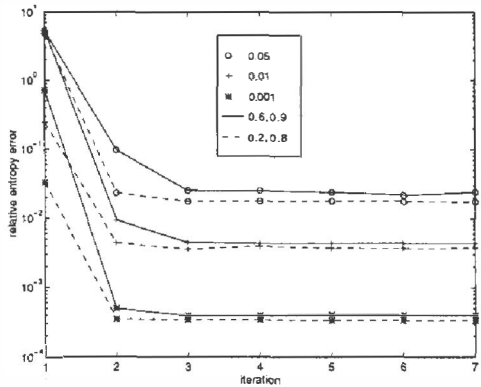

Figure 15: The relative entropy error as a function of the iteration number and the precision parameter $\delta$. Evidence is $o_1 = 0.6$, $o_2 = 0.9$ (solid line) and $o_1 = 0.2$, $o_2 = 0.8$ (dashed line). $o_3$ is always *true*.

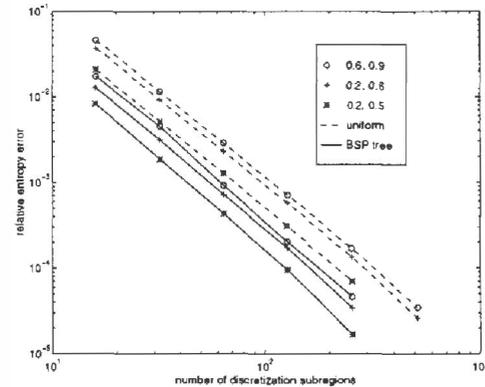

Figure 16: The relative entropy error as a function of the number of discretization subregions and evidence. Evidence is $o_1 = 0.6$, $o_2 = 0.9$ (circles), $o_1 = 0.2$, $o_2 = 0.8$ (pluses), and $o_1 = 0.2$, $o_2 = 0.5$ (stars). $o_3$ is *always true*.

### Acknowledgments

This research was supported through the generosity of the Powell foundation, by ONR grant N00014-96-1-0718, and by ARO under the MURI program "Integrated Approach to Intelligent Systems", grant number DAAH04-96-1-0341.

### References

Alag, S. and Agogino, A. M. (1996). Inference using message propagation and topology transformation vector gaussian continuous networks. *Proceedings of the Twelfth UAI Conference*, pages 20 – 27. Morgan Kaufmann.

Breiman, L., Friedman, J. H., Olshen, R. A., and Stone, C. J. (1984). *Classification and regression trees*. Wadsworth International Group, Belmont, CA.

Cover, T. and Thomas, J. (1991). *Elements of Information Theory*. John Wiley & Sons, Chichester, UK.

Dechter, R. (1996). Bucket elimination: A unifying framework for probabilistic inference. In *Proceedings of the Twelfth UAI Conference*, pages 211 – 219. Morgan Kaufmann.

Driver, E. and Morrel, D. (1995). Implementation of continuous bayesian networks using sums of weighted gaussians. In *Proceedings of the Eleventh UAI Conference*, pages 134 – 140. Morgan Kaufmann.

Lauritzen, S. L. (1992). Propagation of probabilities, means, and variances in mixed graphical association models. *JASA*, 87(420):1089 – 1108.

Lauritzen, S. L. and Spiegelhalter, D. J. (1988). Local computations with probabilities on graphical structures and their application to expert systems. *Journal of the Royal Statistical Society*, B 50:253 – 258.

Lauritzen, S. L. and Wermuth, N. (1989). Graphical models for association between variables, some of which are qualitative and some quantitative. *The Annals of Statistics*, 17(1):31 –57.

Li, Z. and D'Ambrosio, B. (1994). Efficient inference in Bayes networks as a combinatorial optimization problem. *International Journal of Approximate Reasoning*, 11(1):55 – 81.

Moore, A. W. (1991). Variable resolution dynamic programming: Efficiently learning action maps in multivariate real-valued state-spaces. In *Machine Learning: Proceedings of the Eighth International Workshop*, pages 333 – 337.

Olesen, K. G. (1993). Causal probabilistic networks with both discrete and continuous varibales. *IEEE PAMI*, 15(3):275 – 279.

Pearl, J. (1988). *Probabilistic Reasoning in Intelligent Systems: Networks of Plausible Inference*. Morgan Kaufmann.

Samet, H. and Webber, R. (1988). Hierarchical data structures and algorithms for computer graphics. I. Fundamentals. *IEEE Computer Graphics and Applications*, 8(3):48 – 68.

Wellman, M. P. and Liu, C.-L. (1994). State-space abstraction for anytime evaluation of probabilistic networks. In *Proceedings of the Tenth UAI Conference*, pages 567 – 574. Morgan Kaufmann.